\documentclass{article} 
\usepackage{colm2024_conference}
\usepackage{microtype}
\usepackage{hyperref}
\usepackage{url}
\usepackage{booktabs}
\usepackage{CJKutf8}
\usepackage{graphicx}
\usepackage{multicol}
\usepackage{multirow}
\usepackage{array}
\usepackage{amsmath}
\usepackage{booktabs}
\usepackage{enumitem}
\usepackage{wrapfig}
\usepackage{algorithm}
\usepackage{algpseudocode}
\usepackage{microtype}
\usepackage{amsmath}
\usepackage{colortbl}
\usepackage[utf8]{inputenc}
\definecolor{lightgray}{rgb}{0.9,0.9,0.9}
\usepackage{caption}
\usepackage{subcaption}
\usepackage{xcolor}
\usepackage{setspace}
\usepackage{url}
\usepackage{colortbl}
\usepackage{tabularx}
\usepackage{blindtext}
\usepackage{pgfplots}
\pgfplotsset{compat=1.18}
\usepackage{tikz}
\usetikzlibrary{er,positioning,bayesnet}
\usepackage{makecell}
\usepackage{tipa}
\usepackage{siunitx}
\usepackage{nicefrac}
\usepackage{tocloft}
\usepackage{listings}
\usepackage{tcolorbox}
\usepackage{xltabular}
\usepackage{adjustbox}
\usepackage{xurl}
\usepackage{setspace}
\usepackage{lipsum}
\usepackage{longtable}
\usepackage{arydshln}
\usepackage[table]{xcolor}

\newcommand{\linkicon}[2]{%
  \raisebox{-0.2em}{\includegraphics[height=1.1em]{#1}}\quad
  \href{#2}{\texttt{#2}}%
}

\title{
\vspace{-0.0em}
\texorpdfstring{
\begin{tabular}{@{}c@{\hspace{0.2em}}l@{}}
\raisebox{-0.45em}{\includegraphics[height=2em]{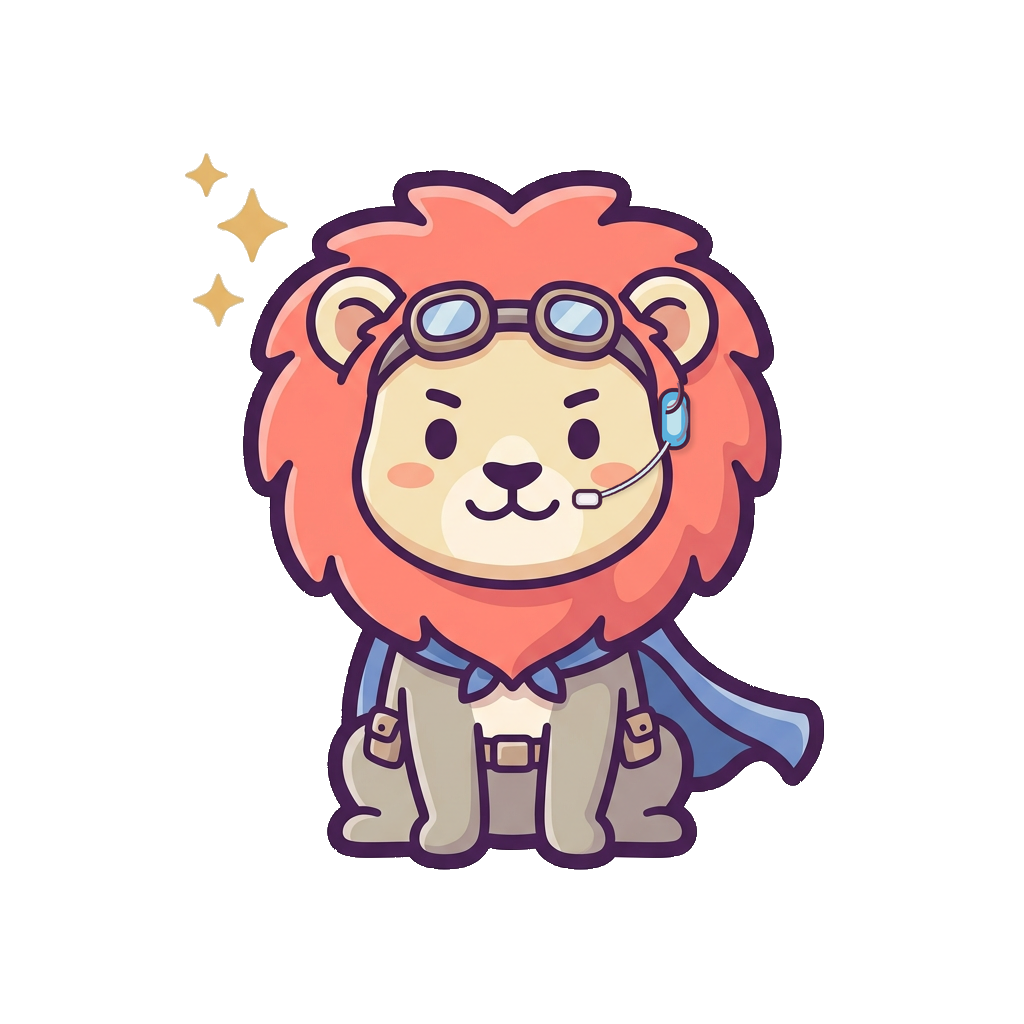}}
&
\begin{tabular}{@{}l@{}}
\Large Hy-MT2: A Family of Fast, Efficient and Powerful\\
\Large Multilingual Translation Models in the Wild
\end{tabular}
\end{tabular}
}{
Hy-MT2: A Family of Fast, Efficient and Powerful Multilingual Translation Models in the Wild
}
\vspace{-0.5em}
}

\author{
	\bf \large Tencent Hunyuan Team
}
\begin{document}
\begin{CJK*}{UTF8}{gbsn}
\maketitle

\begin{abstract}

Hy-MT2 is a family of \textbf{fast-thinking multilingual translation models} designed for complex real-world scenarios. It includes three model sizes: \textbf{1.8B, 7B, and 30B-A3B (MoE)}, all of which support translation among \textbf{33} languages and effectively \textbf{follow translation instructions in multiple languages}.  \textbf{Multi-dimensional evaluations} show that Hy-MT2 delivers outstanding performance across general, real-world business, domain-specific, and instruction-following translation tasks. The 7B and 30B models \textbf{outperform open-source models such as DeepSeek-V4-Pro and Kimi K2.6 in fast-thinking mode}, while the lightweight 1.8B model also \textbf{surpasses mainstream commercial APIs from providers such as Microsoft and Doubao overall}. Moreover, when paired with AngelSlim’s \textbf{1.25-bit} extreme quantization for on-device deployment, the lightweight \textbf{1.8B} model requires only \textbf{440 MB} of storage and achieves a \textbf{1.5×} inference speedup.


\linkicon{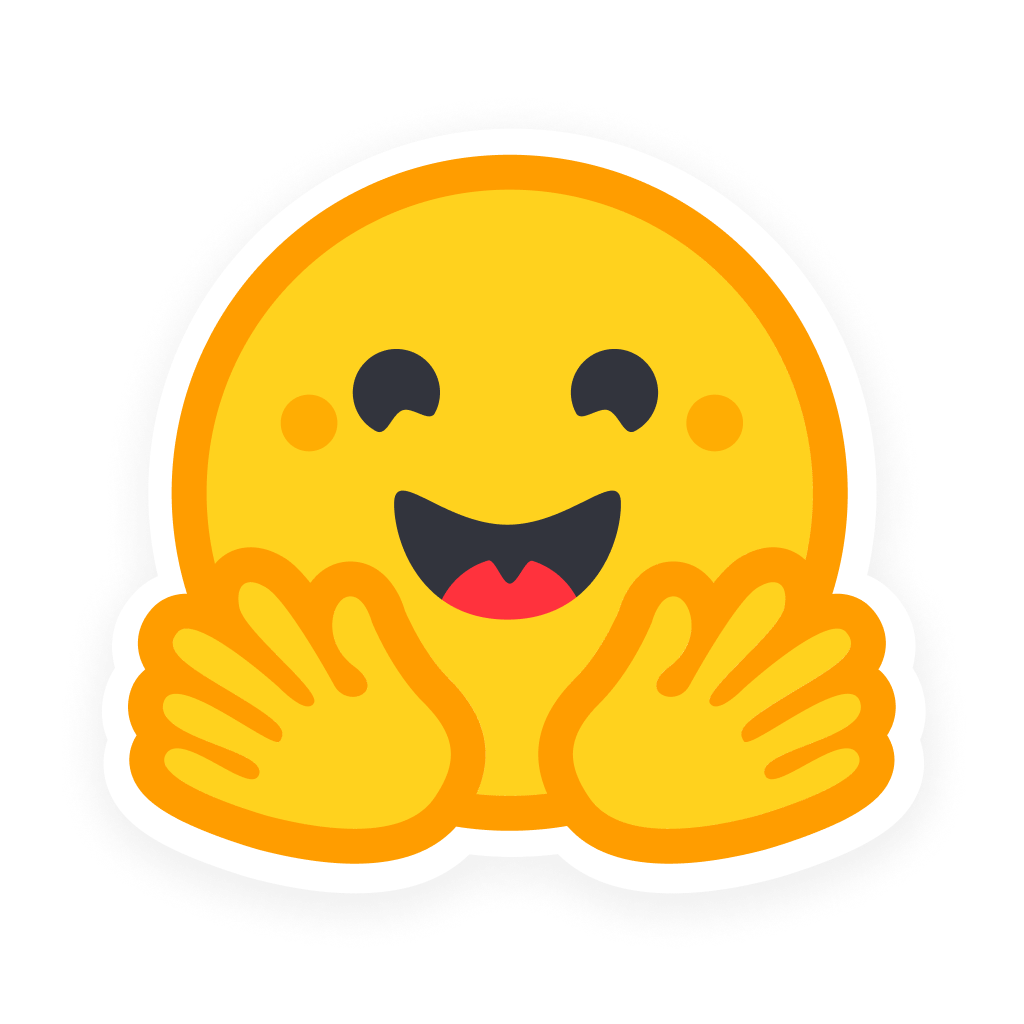}{https://huggingface.co/collections/tencent/hy-mt2}

\linkicon{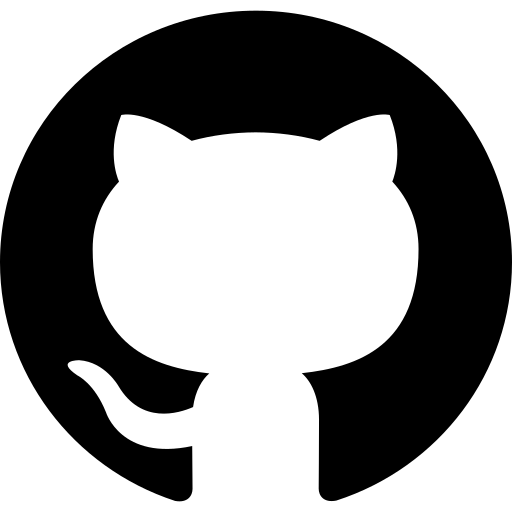}{https://github.com/Tencent-Hunyuan/Hy-MT2}


\end{abstract}

\begin{figure}[h]
\centering
\includegraphics[width=0.9\linewidth]{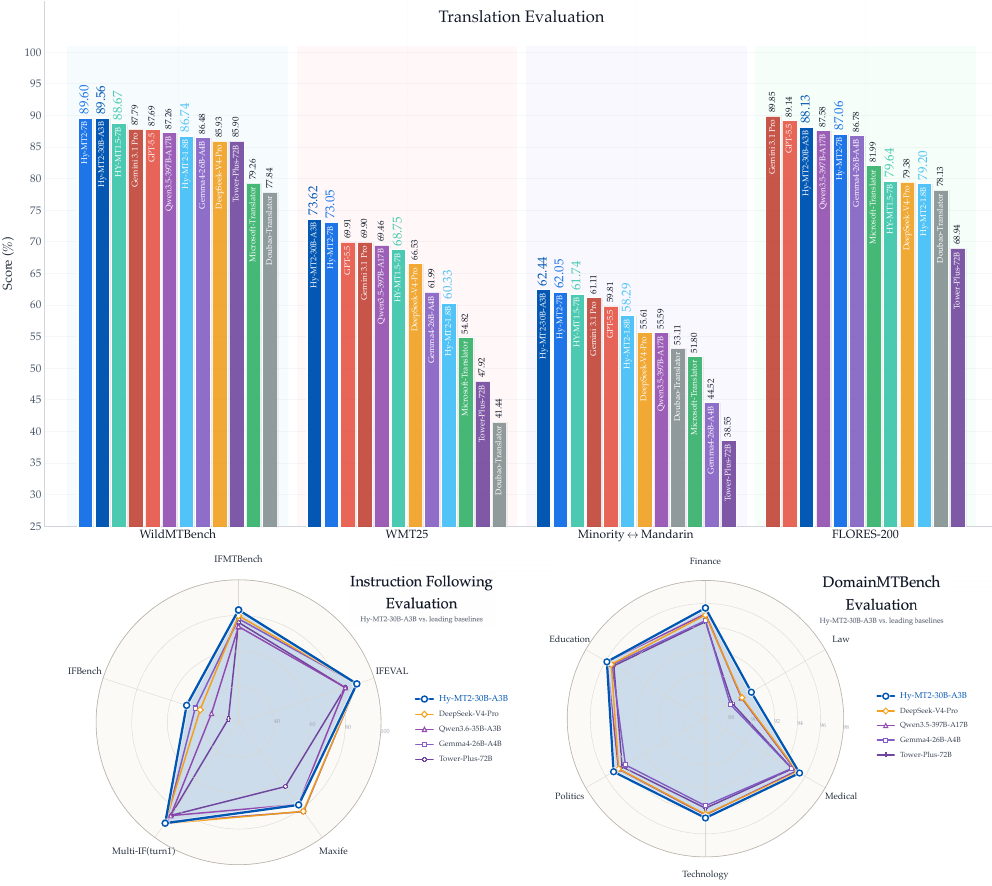}
\caption{Benchmark performance of Hy-MT2 models and state-of-the-art baselines.}
\label{fig:overall}
\end{figure}

\section{From Hy-MT1.5 to Hy-MT2}

After its release, Hy-MT1.5~\citep{hy-mt1.5} attracted broad attention from both the open-source community and real-world business applications. As the model was adopted in more practical translation scenarios, community and business feedback revealed that Hy-MT1.5 still had room for improvement in domain-specific translation, real-world scenario translation, translation instruction following, and efficient on-device deployment. 
Meanwhile, the substantial quality improvements achieved by Hy3-preview further motivated us to leverage it as a strong teacher model to improve the performance of the Hy translation model.
To further address these limitations, we propose the Hy-MT2 model family.

First, domain-specific and real-world scenario translation remain challenging for Hy-MT1.5. Professional domains such as finance, law, and medicine contain a large number of domain-specific terms and established industry translations, placing higher requirements on translation accuracy and consistency. Real-world business scenarios, such as webpages, meetings, and social content, involve more diverse text formats and usage requirements. To address these challenges, Hy-MT2 strengthens translation capabilities for professional domains and real-world application scenarios, enabling the model to better adapt to translation needs across different domains, sources, and text forms.

Second, in practical use, users often impose additional constraints on translation outputs, such as keeping certain words untranslated, controlling the translation style, or producing outputs according to a specified template. In such scenarios, Hy-MT1.5 may ignore constraints or fail to satisfy the specified requirements. To this end, Hy-MT2 enhances multilingual translation instruction understanding and execution, enabling the model to reliably follow user requirements in different languages, including those related to style, format, and other translation constraints, as illustrated in Table~\ref{tab:hymt2_prompt}.

In addition, community feedback indicates that Hy-MT1.5-7B still has a clear gap in translation quality compared with the strongest closed-source models, such as Gemini 3.1 Pro~\citep{gemini_3pro} and GPT-5.5~\citep{singh2026openaigpt5card}. Prior research and model practices suggest that scaling up model size generally helps improve understanding, expression, and instruction-following capabilities in complex translation scenarios. However, representative large-scale translation models, such as TransGemma-27B, mostly adopt dense architectures, leading to high inference costs and making them less suitable for practical service deployment. Therefore, Hy-MT2 introduces a mixture-of-experts architecture and releases Hy-MT2-30B-A3B to achieve a better balance between translation quality and inference efficiency.

Finally, real-world business deployment also exposed the limitations of Hy-MT1.5 in on-device efficiency. The 4-bit quantized version of Hy-MT1.5-1.8B still requires more than 1GB of storage, and its inference speed is insufficient for some low-latency translation scenarios. To address this issue, Hy-MT2 further explores ultra-low-bit quantization and implements 1.25-bit extreme quantization based on Hunyuan’s  AngelSlim technology. This version requires only about 440MB of storage for deployment and achieves a 1.5× inference speedup over the 4-bit quantized Hy-MT1.5 on Apple A15, significantly reducing on-device deployment costs while improving inference efficiency.

Overall, Hy-MT2 systematically addresses the limitations of Hy-MT1.5 in domain-specific translation, real-world scenario translation, translation instruction following, the performance gap with the strongest closed-source models, and efficient on-device deployment. It establishes a high-quality, efficient, and multi-capability multilingual translation model family that is better suited for real-world applications.

\begin{table}[!h]
\centering
\caption{Instruction examples for Hy-MT2 translation tasks in Chinese and English.}
\label{tab:hymt2_prompt}
\scriptsize
\setlength{\tabcolsep}{6pt}
\renewcommand{\arraystretch}{1.25}
\begin{tabularx}{\textwidth}{>{\raggedright\arraybackslash}p{2.2cm}X>{\raggedright\arraybackslash}p{7cm}}
\toprule
\textbf{Type} & \textbf{Chinese prompt} & \textbf{English prompt} \\ 
\midrule

\textbf{Default Translation} & 
将以下文本翻译为\textcolor{blue}{\{target\_lang\}}，注意\textcolor{orange}{只需要输出翻译后的结果，不要额外解释}：\newline\newline 
\textcolor{blue}{\{source\_text\}} & 
Translate the following text into \textcolor{blue}{\{target\_lang\}}. Note that you should \textcolor{orange}{only output the translated result without any additional explanation}:\newline\newline 
\textcolor{blue}{\{source\_text\}} \\ 
\midrule

\textbf{Terminology} & 
\textcolor{purple}{参考下面的翻译：}\newline 
\textcolor{blue}{\{text\}} 翻译成 \textcolor{blue}{\{text\}}\newline 
\textcolor{blue}{\{text\}} 翻译成 \textcolor{blue}{\{text\}}\newline 
\textcolor{blue}{\{text\}} 翻译成 \textcolor{blue}{\{text\}}\newline 
将以下文本翻译为 \textcolor{blue}{\{target\_lang\}}，注意\textcolor{orange}{只需要输出翻译后的结果，不要额外解释}：\newline\newline 
\textcolor{blue}{\{source\_text\}} & 
\textcolor{purple}{Reference the following translations:}\newline 
\textcolor{blue}{\{text\}} translates to \textcolor{blue}{\{text\}}\newline 
\textcolor{blue}{\{text\}} translates to \textcolor{blue}{\{text\}}\newline 
\textcolor{blue}{\{text\}} translates to \textcolor{blue}{\{text\}}\newline\newline 
Translate the following text into \textcolor{blue}{\{target\_lang\}}. Note that you must \textcolor{orange}{ONLY output the translated result without any additional explanation}:\newline\newline 
\textcolor{blue}{\{source\_text\}} \\ 
\midrule

\textbf{Style} & 
请将以下文本翻译为\textcolor{blue}{\{\{target\_lang\}\}}。\newline 
注意翻译的风格要严格符合【\textcolor{orange}{\{\{target\_style\}\}}】\newline\newline 
\textcolor{blue}{\{\{source\_text\}\}} & 
Please translate the following text into \textcolor{blue}{\{\{target\_lang\}\}}. Note that the translation style must strictly conform to [\textcolor{orange}{\{\{target\_style\}\}}]:\newline\newline 
\textcolor{blue}{\{\{source\_text\}\}} \\ 
\midrule

\textbf{Personalization} & 
\textcolor{purple}{【待翻译文本】}\newline 
\textcolor{blue}{\{source\_text\}}\newline\newline 
\textcolor{purple}{【翻译任务】}\newline 
1、\textcolor{orange}{\{user\_preferences\}}\newline 
2、\textcolor{orange}{\{user\_preferences\}}\newline 
3、……\newline 
4、将【待翻译文本】翻译为\textcolor{blue}{\{target\_lang\}}。 & 
\textcolor{purple}{[Source Text]}\newline 
\textcolor{blue}{\{source\_text\}}\newline\newline 
\textcolor{purple}{[Translation Tasks]}\newline 
1. \textcolor{orange}{\{user\_preferences\}}\newline 
2. \textcolor{orange}{\{user\_preferences\}}\newline 
3. ...\newline 
4. Translate the [Source Text] into \textcolor{blue}{\{target\_lang\}}. \\ 
\midrule

\textbf{Delimiters} & 
请将以下文本准确翻译为\textcolor{blue}{\{\{target\_lang\}\}}。\newline 
你必须在译文中\textcolor{orange}{保留等量的分隔符，绝对不可遗漏、转义或翻译该符号，并注意分隔符的位置}。\newline\newline 
\textcolor{blue}{\{\{source\_text\}\}} & 
Please accurately translate the following text into \textcolor{blue}{\{\{target\_lang\}\}}.\newline 
You must \textcolor{orange}{retain the exact same number of delimiters in the translation. Strictly do not omit, escape, or translate these symbols, and pay close attention to their placement}.\newline\newline 
\textcolor{blue}{\{\{source\_text\}\}} \\ 
\midrule

\textbf{Structured Data 1} & 
\textcolor{purple}{\# 任务目标}\newline 
将下方\textcolor{blue}{\{\{source\_text\}\}}中的\textcolor{blue}{\{\{format\_type\}\}}格式数据翻译为\textcolor{blue}{\{\{target\_lang\}\}}。\newline\newline 
\textcolor{purple}{\# 严格约束}\newline 
1. \textcolor{orange}{**结构锁定**}: 绝对保持原有的\textcolor{blue}{\{\{format\_type\}\}}数据结构、缩进和层级完全不变。\newline 
2. \textcolor{orange}{**选择性翻译**}: 仅翻译面向用户展示的可见文本内容。\newline 
3. \textcolor{orange}{**禁止修改**}: **严禁**翻译或更改任何代码标签、键名 (Key)、变量占位符 (如 `\textcolor{blue}{\{\{var\}\}}`, `\textcolor{blue}{\$\{var\}}`, `\textcolor{blue}{\%s}`, `\textcolor{blue}{\%d}` 等) 或代码属性。\newline\newline 
\textcolor{purple}{\# 数据输入}\newline 
\textcolor{blue}{\{\{source\_text\}\}} & 
\textcolor{purple}{\#\#\# Task}\newline 
Translate the user-facing text within the following \textcolor{blue}{\{\{format\_type\}\}} data into \textcolor{blue}{\{\{target\_lang\}\}}.\newline\newline 
\textcolor{purple}{\#\#\# Strict Rules}\newline 
1. \textcolor{orange}{**Structure Preservation:**} You MUST preserve the original \textcolor{blue}{\{\{format\_type\}\}} data structure, nesting, hierarchy, and indentation exactly as they are.\newline 
2. \textcolor{orange}{**Selective Translation:**} Translate ONLY the visible, user-facing text content/values.\newline 
3. \textcolor{orange}{**Strict Non-Translation:**} NEVER translate or alter code tags, keys, properties, object names, or variable placeholders. Leave them exactly in their original English/code form.\newline\newline 
\textcolor{purple}{\#\#\# Source Data}\newline 
\textcolor{blue}{\{\{source\_text\}\}} \\ 
\midrule

\textbf{Structured Data 2} & 
\textcolor{purple}{【背景信息】}\newline 
\textcolor{blue}{\{\{background\_text\}\}}\newline\newline 
请结合背景信息将以下文本翻译为\textcolor{blue}{\{\{target\_lang\}\}}。\newline\newline 
\textcolor{purple}{【待翻译文本】}\newline 
\textcolor{blue}{\{\{source\_text\}\}} & 
\textcolor{purple}{[Background Information]}\newline 
\textcolor{blue}{\{\{background\_text\}\}}\newline\newline 
Please translate the following text into \textcolor{blue}{\{\{target\_lang\}\}}, taking the provided background information into consideration.\newline\newline 
\textcolor{purple}{[Source Text]}\newline 
\textcolor{blue}{\{\{source\_text\}\}} \\ 
\bottomrule
\end{tabularx}
\vspace{0.5em}
\begin{minipage}{0.98\textwidth}
\scriptsize
\textit{Notes.} This table shows representative instruction templates in Chinese and English. 
Additional multilingual instruction examples are provided in Section~\ref{sec:case_study}.
\end{minipage}
\end{table}

\section{Methodology}

\begin{figure}[h]
\centering
\includegraphics[width=1\linewidth]{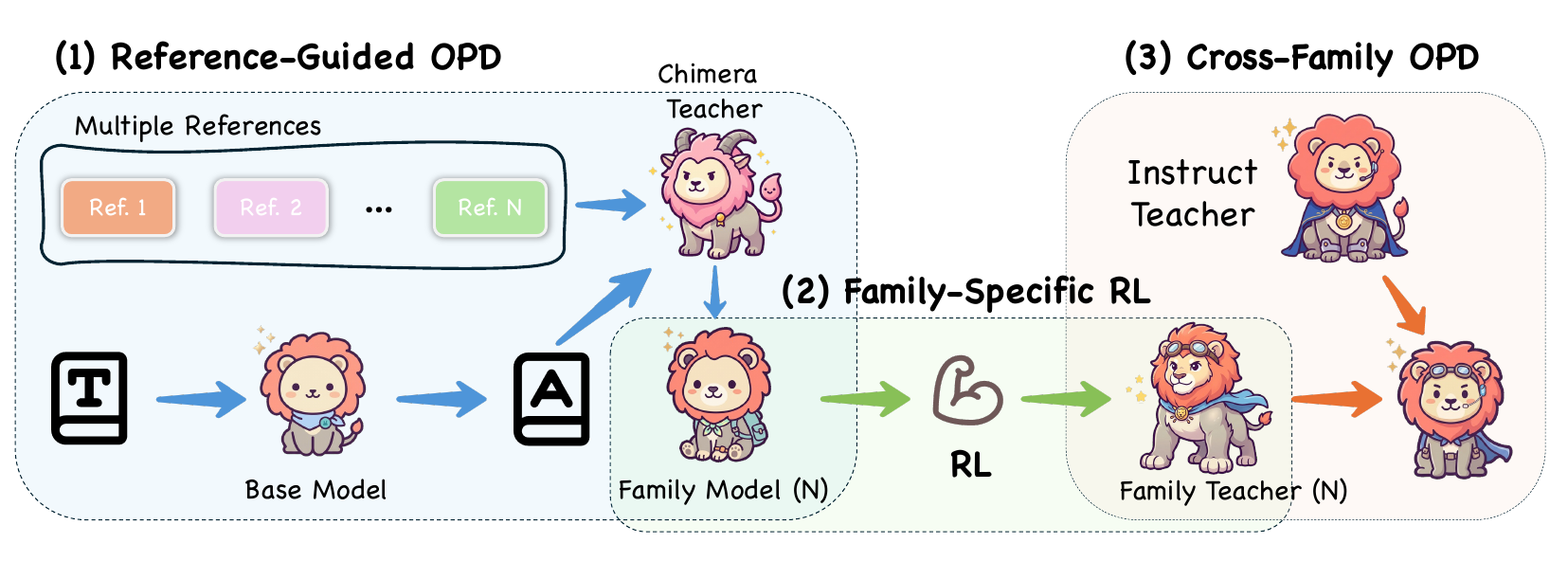}
\caption{Family-Centric Post-training pipline of Hy-MT2. }
\label{fig:framework}
\end{figure}

This section introduces the overall methodology of Hy-MT2. Designed for multilingual machine translation, Hy-MT2 follows a staged pipeline consisting of \textbf{MT-oriented Mid-training} (Section~\ref{sec:mid_train}), \textbf{Family-Centric Post-training} (FCPT；Section~\ref{sec:fcpt}), and model \textbf{Quantization} (Section~\ref{sec:quantization}). Specifically, we start from a general Hy-series Pretraining Model and perform MT-oriented Mid-training to obtain a unified model with fundamental translation capabilities. The model is then further optimized through FCPT.  As shown in Figure~\ref{fig:framework}, FCPT consists of three key processes: \textbf{Reference-Guided On Policy Distillation }(RG-OPD; Section~\ref{sec:rgopd}), \textbf{Family-specific RL Training} (Section~\ref{sec:rl}), and \textbf{Cross-family On Policy Distillation} (Cross-family OPD; Section~\ref{sec:cfopd}). 
The first two processes organize training around language families and construct multiple family-specific strong teachers; Cross-family OPD then transfers their capabilities into a unified student model while incorporating general instruction-following data to preserve the model's instruction-following ability beyond translation.

\subsection{MT-oriented Mid-training}\label{sec:mid_train}

In the \textbf{MT-oriented Mid-training} stage, we start from the Hy-series Pretraining Model and continue training it on approximately 1T tokens of large-scale multilingual translation-related data. This stage aims to strengthen the model's translation capability and provide a unified foundation for the subsequent Family-Centric Post-training.

Specifically, the training data is organized along two dimensions:
\begin{itemize}
    \item \textbf{Data format}: We use both multilingual monolingual corpus and parallel translation corpus to help the model capture linguistic characteristics across different languages and strengthen cross-lingual semantic mapping and source-target alignment.
    \item \textbf{Scenario coverage}: The data covers general translation, domain-specific translation, real-world scenarios, and instruction-following examples, improving translation quality, domain adaptation, practical translation robustness, and the ability to follow translation-related instructions.
\end{itemize}

The output of this stage is an \textbf{MT-oriented Mid-trained Model}, which serves as the unified starting point for \textbf{FCPT}.

\subsection{Family-Centric Post-training}\label{sec:fcpt}

Instead of directly mixing data from all language families, FCPT divides training into multiple family branches, covering diverse language groups, e.g., Western European, East Asian, and Middle Eastern right-to-left languages.  Within each branch, we incorporate general translation data, domain-specific translation data, real-world business scenario data, and translation instruction-following data to construct a family-specific teacher. 
This family-centric design allows each teacher to learn under a more consistent language distribution, reducing interference across different language families.

\subsubsection{Reference-Guided On Policy Distillation}\label{sec:rgopd}

\textbf{Reference-Guided On Policy Distillation} is the first stage of FCPT. In this stage, we perform On policy distillation separately on each family branch, aiming to obtain a family-specific translation policy that better captures the linguistic characteristics and translation preferences of the corresponding language family. The resulting model further serves as a stronger initialization for subsequent Family-specific RL Training.

The core of RG-OPD is the construction of a stronger Chimera Teacher. Unlike conventional distillation methods that rely on a single teacher model, Chimera\footnote{The name \textit{Chimera} is inspired by the mythological creature composed of multiple animals. In our setting, it refers to a teacher signal constructed by fusing multiple reference sources. We implement Chimera Teacher based on Hy3-Preview.}. Teacher does not require training an additional large-scale translation-specialized teacher model. For each source sentence, it integrates candidate translations generated by multiple Hy-series reference models with the original dataset label. Although not all labels are manually annotated, they still serve as useful reference signals. By fusing these multi-source references, Chimera Teacher provides richer scoring signals, helps introduce greater diversity into the distillation process, and constructs a stronger supervision signal for On policy distillation.

Specifically, given a source sentence $x$ and its reference set $\mathcal{R}(x)$, where $\mathcal{R}(x)$ consists of multiple candidate reference sources, the student model in the current family branch first generates a translation $y$ based on its current policy $\pi_{\theta}$. Chimera Teacher then evaluates the student output based on the multi-source reference set $\mathcal{R}(x)$, and produces a teacher policy or target distribution $\pi_T(\cdot \mid x, \mathcal{R}(x))$. The student model is optimized by minimizing the forward KL divergence from the teacher policy to the student policy. The training objective can be written as:
\begin{equation}
\mathcal{L}_{\text{RG-OPD}} = D_{\text{KL}} \left( \pi_T(\cdot \mid x, \mathcal{R}(x)) \parallel \pi_{\theta}(\cdot \mid x) \right).
\end{equation}
Here, $\pi_T$ denotes the distillation target distribution constructed by Chimera Teacher, and $\pi_{\theta}$ denotes the output policy of the current student model. We adopt forward KL divergence as the distillation objective, enabling the student model to learn the fused translation preference from Chimera Teacher in an online manner and gradually improve its translation policy for the corresponding language family.

After \textbf{RG-OPD}, each family branch obtains a family-specific student model that has been adapted to the translation preferences and expression patterns of the corresponding language family. These models are then used as the initialization for subsequent \textbf{Family-specific RL Training}.

\subsubsection{Family-specific RL Training}\label{sec:rl}

In \textbf{Family-specific RL Training}, each family branch is further optimized through Group Relative Policy Optimization (GRPO)~\citep{grpo}, using the model obtained from RG-OPD as initialization. To provide more fine-grained and rigorous reward signals, we introduce a hybrid evaluation system combining a rule-based pre-filter with an LLM-based Multidimensional Quality Metrics (MQM) judge~\citep{14-MQM,21-wmt-mqm}.

\textbf{Rule-based Pre-filtering} \
Before passing translations to the LLM evaluator, a rule-based filter is applied to intercept critical text degradation. Translations exhibiting severe repetition or mixed languages are immediately assigned a reward of 0. This ensures early penalization of degenerated outputs and avoids unnecessary LLM computation.

\textbf{LLM Judge Evaluation System} \
For translations that pass the pre-filter, the LLM-based judge evaluates the text based on a 5-dimensional error typology rather than assigning a holistic score. The dimensions are:
\begin{itemize}
    \item \textbf{Terminology}: Identifies terminology errors, inconsistencies with terminological resources, or inconsistent usage throughout the text.
    \item \textbf{Accuracy}: Detects mistranslations, over-translation, under-translation, added or omitted content, unwarranted translations, missed translations, and instances of mixed languages.
    \item \textbf{Linguistic Conventions}: Checks for grammatical errors, punctuation errors, spelling mistakes, unintelligibility, discourse convention errors, and locale convention violations.
    \item \textbf{Style}: Assesses inconsistencies with external references, incorrect language register, obscure expressions, unnatural phrasing, and stylistic inconsistencies.
    \item \textbf{Instruction Following}: Evaluates adherence to task constraints, flagging wrong languages, unexecuted translation tasks, and failures to follow terminology, formatting, style, or context guidelines.
\end{itemize}

\textbf{Scoring Rules} \
The evaluation starts from a base score of 100. The LLM judge identifies translation errors and applies deductions according to their severity. Fatal errors, such as using the wrong language or failing to execute the translation task, directly result in an overall score of 0. Major errors incur a deduction of 10--20 points per instance, while minor errors incur a deduction of 2--5 points per instance. The overall score $S_{\text{overall}}$ is obtained by subtracting all error deductions from the base score, with a lower bound of 0.


\textbf{Length Penalty and Final Reward Calculation} \
To prevent the model from exploiting the reward system by generating pathologically short, truncated, or excessively long and redundant sentences, we introduce a length penalty. Given a source sentence $x$, a ground-truth translation of length $L_{gt}$, and a model-generated translation $y$ of length $L_y$, the length penalty $P_{\text{len}}$ is computed as:

\begin{equation}
P_{\text{len}}=\min\left(0.5\times\frac{|L_{gt}-L_y|}{L_{gt}},0.5\right).
\end{equation}


The final reward $r(x,y)$ is calculated by normalizing the overall MQM score to the range $[0,1]$, subtracting the length penalty $P_{\text{len}}$, and clipping the result at 0.

After this stage, each family branch produces a strong expert, which is then used as a strong teacher in the subsequent \textbf{Cross-family OPD} stage.

\subsubsection{Cross-family On Policy Distillation}\label{sec:cfopd}

\textbf{Cross-family On Policy Distillation }is the final training process of FCPT, aiming to transfer the language-family-specific translation capabilities learned by multiple family-specific strong teachers into a unified student model.  Meanwhile, to improve the model’s instruction-following ability, we introduce general instruction-following data in this stage and use the Hy Instruct model as the corresponding instruction teacher to provide distillation signals. Therefore, Cross-family OPD can be viewed as a unified multi-teacher distillation process.

Specifically, given an input sample $x$, we denote its teacher policy as $\pi_T(\cdot \mid x)$ and the output policy of the unified student model as $\pi_{\theta}(\cdot \mid x)$. In this stage, we adopt reverse KL divergence as the distillation objective, which is defined as:
\begin{equation}
\mathcal{L}_{\text{Cross-OPD}} = D_{\text{KL}} \Big( \pi_{\theta}(\cdot \mid x) \parallel \pi_{T_{\tau(x)}}(\cdot \mid x) \Big),
\end{equation}
where $\pi_{\theta}(\cdot \mid x)$ denotes the output policy of the student model initialized from MT-oriented Mid-training, and $\pi_{T_{\tau(x)}}(\cdot \mid x)$ denotes the output policy of the selected teacher. For translation samples, $\tau(x)$ selects the corresponding family-specific strong teacher according to the language family of $x$; for general instruction-following samples, $\tau(x)$ selects the Hy Instruct teacher.

After \textbf{Cross-family On Policy Distillation}, we obtain the final \textbf{Hy-MT2} series models. To facilitate efficient deployment, these models are further processed in the quantization stage.

\subsection{Quantization}\label{sec:quantization}
To accommodate deployment requirements under varying resource constraints, we perform model quantization on the obtained Hy-MT2 model series, offering a diverse suite of precision variants including FP16, 8-bit, 4-bit, 2-bit, and 1.25-bit. 

For the 8-bit and 4-bit variants, we predominantly adopt a post-training quantization (PTQ) pipeline. Without retraining the model, this approach estimates the distribution of model weights or activations using a small set of calibration data, thereby reducing storage and computational overheads. Specifically, 8-bit quantization employs a higher-precision low-bit representation to minimize performance degradation, whereas 4-bit quantization further compresses weight representations and mitigates quantization errors through dedicated calibration strategies. 

For the 2-bit version, we employ the ultra-low-bit quantization-aware training (QAT) scheme from AngelSlim~\cite{angelslim2026} framework. Compared with PTQ, 2-bit quantization imposes more stringent representational constraints. Consequently, it requires explicit simulation of low-bit quantization behaviors during the training process, allowing model weights to progressively adapt to low-precision representations. The 2-bit scheme utilizes Stretched Elastic Quantization (SEQ)~\cite{liu2026paretoq}, quantizing weight into {-1.5, -0.5, +0.5, +1.5}. By optimizing the quantization mapping and scaling factors, this approach enhances the model's stability and capability for performance recovery under 2-bit constraints.

For extreme compression scenarios, we further implement Sherry~\cite{huang2026sherry}, a 1.25-bit sparse ternary quantization method. Sherry quantizes model weights into a ternary space of $\{-1, 0, +1\}$ and introduces a 3:4 fine-grained sparsity pattern, where constraining every 4-weight block to contain exactly one zero and three sign-only weights. This structure enables packing four weights into 5 bits, thereby achieving a regularized 1.25-bit representation while maintaining a hardware alignment that is more favorable for Single Instruction Multiple Data (SIMD) computing patterns. Compared to conventional 2-bit packing or 1.67-bit irregular packing, Sherry achieves a superior balance between compression ratio and inference efficiency.

For 1.25-bit and 2-bit QAT, we adopt a distillation-based training strategy. Specifically, the low-precision model serves as the student, while the high-precision model acts as the teacher. The overall objective combines the standard language modeling (LM) loss with a KL-divergence-based distillation loss. For the KL component, inspired by~\cite{cakld}, we incorporate both forward and backward KL divergences. Unlike prior approaches that use a fixed, manually selected weighting coefficient, we compute the teacher model’s confidence score for each token and dynamically adjust the relative weights of the forward and backward KL terms accordingly.

Ultimately, we obtain the quantized Hy-MT2 model series spanning FP16, 8-bit, 4-bit, 2-bit, and 1.25-bit precisions. Armed with these diverse precision variants, Hy-MT2 can be flexibly deployed across a wide range of scenarios, including high-accuracy service serving, low-resource device inference, and extreme edge-side compression.

\section{Experiments}

\subsection{Benchmarks}

To comprehensively evaluate the translation capabilities of Hy-MT2, we construct an evaluation suite from four perspectives: general translation, real-world business scenario translation, domain-specific translation, and translation instruction following.

\textbf{General translation evaluation.} \ We use FLORES-200~\citep{flores-200}, WMT25~\citep{kocmi-etal-2025-findings}, and the Mandarin$\Leftrightarrow$Minority Testset to evaluate general translation capabilities. FLORES-200 covers 1,056 translation directions across 33 languages. WMT25 adopts the human evaluation sets from WMT25 and covers 12 translation directions. The Mandarin$\Leftrightarrow$Minority Testset focuses on bidirectional translation between Mandarin Chinese and minority languages.

\textbf{Real-world business scenario evaluation.} We construct WildMTBench to assess model performance on practical business inputs. The dataset covers six types of scenarios, including webpages, meetings, books, social content, news, and documents, with 2,000 samples in total. It focuses on evaluating model robustness and adaptability to diverse text forms, real-world input distributions, and complex application requirements.

\textbf{Domain-specific translation evaluation.} \ We construct DomainMTBench to assess translation quality in professional domains. The dataset covers six domains: finance, law, politics, technology, medicine, and education. The data are collected from open-source corpora and processed through cleaning, filtering, domain classification, and human translation annotation, resulting in 24,000 samples. This benchmark focuses on evaluating the model's ability to handle domain terminology, specialized expressions, and established industry translations.

\textbf{Instruction-following evaluation.} \ We construct IFMTBench to evaluate translation-specific instruction following in multilingual settings. It contains 7,344 high-quality human-aligned samples with instructions in Chinese, German, Japanese, French, English, Spanish, and Korean, covering industrial translation constraints such as terminology, format, and style. It includes 4,506 single-constraint and 2,838 multi-constraint samples, evaluating both basic constraint execution and robustness under complex instruction combinations.
In addition, we use IFBench~\citep{IFBench}, IFEval~\citep{IFEval}, MaXIFE~\citep{maxife}, and Multi-IF~\citep{Multi_IF} to assess general instruction-following capabilities. IFEval is a verifiable instruction-following benchmark for large language models, containing around 500 prompts. IFBench focuses on model generalization to diverse and unseen verifiable constraints. MaXIFE evaluates multilingual and cross-lingual instruction following across 23 languages. Multi-IF focuses on instruction following in multi-turn and multilingual interactions.

\begin{table*}[t]
\centering
\caption{Performance comparison on general translation benchmarks.}
\label{tab:general_translation_results_compact}
\scriptsize
\renewcommand{\arraystretch}{1.23}
\setlength{\tabcolsep}{4pt}
\resizebox{\textwidth}{!}{
\begin{tabular}{lccccc}
\toprule
\multirow{2}{*}{\textbf{Models}} 
& \multicolumn{3}{c}{\textbf{FLORES-200}} 
& \multirow{2}{*}{\textbf{WMT25}} 
& \multirow{2}{*}{\textbf{Mand.$\Leftrightarrow$Min.}} \\
\cmidrule(lr){2-4}
& \textbf{ZH $\Leftrightarrow$ XX} 
& \textbf{EN $\Leftrightarrow$ XX} 
& \textbf{XX $\Leftrightarrow$ XX} 
& & \\
\midrule

{\color{blue}Gemini 3.1 Pro$^{\mathrm{T}}$} 
& \colorbox{cyan!20}{90.30} / 78.96 / \colorbox{cyan!20}{92.14} 
& \colorbox{cyan!20}{94.42} / \colorbox{cyan!20}{88.38} / 92.68 
& \colorbox{cyan!20}{88.74} / \colorbox{cyan!20}{77.60} / \colorbox{cyan!20}{90.97} 
& \colorbox{cyan!20}{57.58} / 69.06 / 82.23 
& \colorbox{cyan!20}{61.11} / 53.50 / \colorbox{cyan!20}{79.70} \\

{\color{blue}GPT-5.5$^{\mathrm{T}}$} 
& 89.94 / 78.95 / 91.98 
& 94.16 / 88.33 / \colorbox{cyan!20}{92.76} 
& 88.44 / 77.36 / 90.93 
& 56.68 / \colorbox{cyan!20}{69.31} / 83.29 
& 60.87 / \colorbox{cyan!20}{55.18} / 79.68 \\

{\color{blue}GPT-5.5} 
& 89.60 / \colorbox{cyan!20}{78.96} / 91.65 
& 93.98 / 88.17 / 92.54 
& 87.92 / 77.20 / 90.37 
& 56.41 / 69.15 / \colorbox{cyan!20}{83.41} 
& 59.81 / 53.32 / 75.16 \\

{\color{blue}DeepSeek-V4-Pro$^{\mathrm{T}}$} 
& 88.86 / 78.13 / 90.97 
& 93.22 / 87.83 / 91.73 
& 86.81 / 76.66 / 89.86 
& 54.83 / 68.25 / 81.99 
& 56.31 / 51.24 / 69.48 \\

{\color{blue}DeepSeek-V4-Pro} 
& 88.60 / 78.11 / 90.15 
& 93.02 / 87.71 / 91.39 
& 83.29 / 67.64 / 75.46 
& 53.97 / 67.58 / 79.09 
& 55.61 / 51.47 / 66.02 \\

{\color{blue}Kimi K2.6$^{\mathrm{T}}$} 
& 88.49 / 77.97 / 90.84 
& 92.95 / 87.72 / 91.28 
& 86.17 / 76.72 / 89.44 
& 54.48 / 68.64 / 81.68 
& 55.98 / {54.76} / 68.73 \\

{\color{blue}Kimi K2.6} 
& 87.68 / 77.55 / 89.85 
& 91.96 / 86.95 / 90.31 
& 84.05 / 75.84 / 87.43 
& 49.76 / 66.08 / 76.84 
& 54.29 / 52.52 / 64.00 \\

{\color{blue}GLM5.1$^{\mathrm{T}}$} 
& 88.78 / 78.28 / {91.38} 
& 93.26 / 87.82 / 91.84 
& 86.57 / 76.84 / 89.92 
& 54.23 / 68.39 / 80.71 
& 57.51 / 54.46 / {71.72} \\

{\color{blue}GLM5.1} 
& 87.09 / 77.60 / 89.44 
& 91.59 / 86.73 / 89.92 
& 83.83 / 75.40 / 87.14 
& 49.71 / 65.41 / 73.70 
& 56.00 / 50.68 / 67.70 \\

{\color{blue}Qwen3.5-397B-A17B$^{\mathrm{T}}$} 
& 87.79 / 77.75 / 90.58 
& 92.38 / 87.19 / 90.65 
& 86.06 / 76.52 / 89.28 
& 54.95 / 68.63 / 81.37 
& 55.97 / 51.89 / 70.21 \\

{\color{blue}Qwen3.5-397B-A17B} 
& 88.50 / 78.64 / 90.66 
& 93.07 / 87.73 / 91.49 
& 86.29 / 76.39 / 88.87 
& 55.79 / 68.80 / 83.14 
& 55.59 / 54.44 / 67.45 \\

{\color{blue}Qwen3.6-35B-A3B$^{\mathrm{T}}$} 
& 87.71 / 77.93 / 90.32 
& 92.32 / 87.55 / 90.49 
& 84.84 / 75.95 / 88.21 
& 50.75 / 65.24 / 73.42 
& 54.07 / 53.37 / 64.44 \\

{\color{blue}Gemma4-31B$^{\mathrm{T}}$} 
& 89.30 / 78.86 / 91.16 
& 93.79 / {88.27} /{91.90} 
& {87.77} / {77.28} /{90.11} 
& 54.48 / 67.12 / 79.59 
& 51.99 / 51.49 / 64.17 \\

{\color{blue}Gemma4-26B-A4B$^{\mathrm{T}}$} 
& 88.68 / 78.68 / 90.74 
& 93.31 / 88.09 / 91.53 
& 86.80 / 77.14 / 89.58 
& 52.13 / 65.40 / 77.73 
& 47.83 / 49.19 / 56.73 \\

{\color{blue}Gemma4-E4B$^{\mathrm{T}}$} 
& 83.49 / 75.94 / 83.96 
& 89.97 / 85.90 / 87.67 
& 81.29 / 74.10 / 83.74 
& 39.70 / 56.48 / 62.34 
& 41.16 / 40.64 / 42.25 \\

{\color{blue}Gemma4-E2B$^{\mathrm{T}}$} 
& 83.21 / 76.18 / 84.21 
& 88.94 / 85.38 / 86.39 
& 79.69 / 73.78 / 82.54 
& 37.07 / 53.51 / 57.21 
& 40.44 / 40.39 / 42.67 \\

\hline

{\color{orange}Qwen3.6-35B-A3B} 
& 86.32 / 78.00 / 87.34 
& 90.87 / 86.52 / 88.77 
& 82.11 / 74.81 / 84.92 
& 51.11 / 66.70 / 75.28 
& 52.12 / \colorbox{orange!20}{51.74} / 58.28 \\

{\color{orange}Gemma4-31B} 
& 88.40 / 78.68 / 89.89 
& 93.30 / 87.96 / \colorbox{orange!20}{91.03} 
& 86.84 / {76.16} / {88.67} 
& 52.49 / 65.66 / 75.75 
& 48.98 / 50.05 / 56.75 \\

{\color{orange}Gemma4-26B-A4B} 
& 87.61 / 78.31 / 88.97 
& 92.80 / 87.71 / 90.65 
& 85.60 / 76.58 / 87.97 
& 49.62 / 64.22 / 74.36 
& 44.52 / 46.14 / 47.70 \\

{\color{orange}Gemma4-E4B} 
& 83.36 / 75.92 / 83.52 
& 89.54 / 85.61 / 86.95 
& 79.96 / 72.91 / 80.97 
& 38.36 / 55.45 / 59.95 
& 41.17 / 40.23 / 43.19 \\

{\color{orange}Gemma4-E2B} 
& 79.78 / 73.19 / 76.46 
& 87.20 / 84.00 / 83.63 
& 75.97 / 71.00 / 76.64 
& 35.60 / 51.83 / 54.35 
& 39.07 / 33.83 / 34.62 \\

{\color{purple}Tower-Plus-72B} 
& 79.69 / 71.82 / 77.86 
& 84.16 / 81.13 / 78.82 
& 70.02 / 65.53 / 67.85 
& 41.00 / 55.54 / 54.85 
& 38.55 / 35.40 / 26.70 \\

{\color{purple}translategemma-27b-it} 
& -- / -- / -- 
& -- / -- / -- 
& -- / -- / -- 
& 58.02 / 66.08 / 81.24 
& -- / -- / -- \\

{\color{purple}Microsoft-Translator} 
& 82.95 / 72.85 / 84.79 
& 89.31 / 85.48 / 87.34 
& 80.13 / 72.42 / 83.85 
& 42.01 / 60.21 / 67.63 
& 51.80 / 41.77 / 55.27 \\

{\color{purple}Doubao-Translator} 
& 80.92 / 71.57 / 82.14 
& 86.77 / 83.49 / 84.46 
& 76.54 / 69.92 / 79.72 
& 33.13 / 54.52 / 49.75 
& 53.11$^*$ / 40.16$^*$ / 66.81$^*$ \\

{\color{purple}iFLYTEK-Translator} 
& 83.00 / 73.96 / -- 
& 87.18 / 83.65 / -- 
& 76.53 / 69.90 / -- 
& 35.10 / 56.15 / -- 
& 59.04 / 44.67 / -- \\

\hdashline

\textbf{HY-MT1.5-1.8B} 
& 83.61 / 76.55 / 76.79 
& 89.42 / 84.11 / 81.35 
& 78.40 / 71.82 / 75.12 
& 53.08 / 61.95 / 63.58 
& 58.06 / 40.84 / 61.66 \\

\textbf{HY-MT1.5-7B} 
& 86.90 / \colorbox{orange!20}{79.24} / 84.35 
& 90.93 / 86.50 / 85.66 
& 80.98 / 73.36 / 78.30 
& 61.59 / 68.85 / 75.91 
& 61.74 / 44.55 / 67.26 \\

\hdashline

\textbf{Hy-MT2-1.8B} 
& 84.26 / 76.43 / 82.71 
& 90.00 / 85.46 / 84.05 
& 79.77 / 73.41 / 78.64 
& 50.30 / 64.59 / 70.36 
& 58.29 / 39.22 / 62.36 \\

\textbf{Hy-MT2-7B} 
& 89.45 / 78.97 / 88.89 
& 93.52 / 88.07 / 89.34 
& 86.89 / 76.03 / 87.23 
& \colorbox{orange!20}{63.86} / \colorbox{orange!20}{71.21} / 82.24 
& 62.05 / 43.60 / 68.93 \\

\textbf{Hy-MT2-30B-A3B} 
& \colorbox{orange!20}{89.83} / {79.03} / \colorbox{orange!20}{90.26} 
& \colorbox{orange!20}{93.85} / \colorbox{orange!20}{88.23} / {90.89} 
& \colorbox{orange!20}{87.47} / \colorbox{orange!20}{76.34} / \colorbox{orange!20}{88.79} 
& 62.89 / 71.08 / \colorbox{orange!20}{84.34} 
& \colorbox{orange!20}{62.44} / 42.37 / \colorbox{orange!20}{69.43} \\

\bottomrule
\end{tabular}
}
\vspace{1mm}

\begin{minipage}{0.98\textwidth}
\tiny
\textit{Notes.} Each cell reports XCOMET-XXL / CometKiwi / GEMBA scores, and all scores are multiplied by 100. 
$^{\mathrm{T}}$ denotes thinking mode. 
In FLORES-200, XX$\Leftrightarrow$XX denotes the average performance over all evaluated translation directions, including ZH$\Leftrightarrow$XX and EN$\Leftrightarrow$XX. 
Mand.$\Leftrightarrow$Min. denotes Mandarin$\Leftrightarrow$Minority translation. 
Values marked with $^*$ are computed only on supported language pairs. 
Values replaced by -- indicate that the model does not support the corresponding test set. 
Baselines are grouped into 
{\color{blue}large-scale models and all Think-mode models}, 
{\color{orange}medium to small-sized general models in non-Think mode}, and 
{\color{purple}translation-specialized models}. 
Our models are shown in bold. The best results among large-scale models and all Think-mode models in each column are highlighted in \colorbox{cyan!20}{blue background}, while the best results among non-thinking-mode, small-to-medium-sized models are highlighted in \colorbox{orange!20}{light orange background}.
\end{minipage}
\end{table*}

\subsection{General Translation Evaluation}

In general translation evaluation, we use three metrics: XCOMET-XXL, CometKiwi, and GEMBA. XCOMET-XXL is a reference-based automatic evaluation metric, CometKiwi is a reference-free evaluation metric, and GEMBA is an LLM-based evaluation metric. The results are reported in Table~\ref{tab:general_translation_results_compact}.

Overall, Hy-MT2 achieves substantial improvements over Hy-MT1.5 across general translation benchmarks. On the XX$\Leftrightarrow$XX setting of FLORES-200, Hy-MT2-1.8B, Hy-MT2-7B, and Hy-MT2-30B-A3B reach 79.77, 86.89, and 87.47, corresponding to 89.9\%, 97.9\%, and 98.6\% of Gemini 3.1 Pro$^{\mathrm{T}}$, respectively. In particular, Hy-MT2-7B and Hy-MT2-30B-A3B outperform strong baselines such as DeepSeek-V4-Pro, Kimi K2.6, Qwen3.5-397B-A17B, and Gemma4-26B-A4B in this setting. Compared with Hy-MT1.5-7B, Hy-MT2-7B improves the XCOMET-XXL score from 80.98 to 86.89, showing a clear gain in overall multilingual translation performance.

On WMT25, Hy-MT2-7B and Hy-MT2-30B-A3B also show strong performance. Hy-MT2-7B achieves 63.86 / 71.21 / 82.24, while Hy-MT2-30B-A3B achieves 62.89 / 71.08 / 84.34. Compared with Hy-MT1.5-7B, whose scores are 61.59 / 68.85 / 75.91, Hy-MT2-7B improves on all three metrics, with a particularly large gain on GEMBA. Hy-MT2-30B-A3B further achieves the best GEMBA score among all compared systems, surpassing Gemini 3.1 Pro$^{\mathrm{T}}$ and GPT-5.5, indicating stronger overall translation quality and readability in challenging WMT settings.

On Mandarin$\Leftrightarrow$Minority translation, Hy-MT2-7B and Hy-MT2-30B-A3B achieve XCOMET-XXL scores of 62.05 and 62.44, respectively, outperforming both Gemini 3.1 Pro and Hy-MT1.5-7B. This suggests that Hy-MT2 preserves strong performance in Mandarin-minority language translation and further improves translation quality in low-resource language scenarios.

For the lightweight setting, Hy-MT2-1.8B shows consistent improvements over Hy-MT1.5-1.8B and remains highly competitive against larger open-source models and commercial translation systems. Despite its small size, it outperforms Tower-Plus-72B and achieves competitive results against Microsoft Translator and Doubao Translator. On WMT25, Hy-MT2-1.8B surpasses both commercial systems across all three metrics, demonstrating a strong efficiency-quality trade-off.

\subsection{Domain-Specific and In-the-Wild Translation Evaluation}

\begin{table*}[t]

\centering

\caption{Performance comparison on DomainMTBench and WildMTBench.}

\label{tab:domain_wild_results}

\scriptsize

\renewcommand{\arraystretch}{1.23}

\setlength{\tabcolsep}{3.5pt}

\resizebox{\textwidth}{!}{

\begin{tabular}{lcccccccc}

\toprule

\multirow{2}{*}{\textbf{Models}} 

& \multicolumn{7}{c}{\textbf{DomainMTBench}} 

& \multirow{2}{*}{\textbf{WildMTBench}} \\

\cmidrule(lr){2-8}

& \textbf{Finance} 

& \textbf{Law} 

& \textbf{Medical} 

& \textbf{Technology} 

& \textbf{Politics} 

& \textbf{Education} 

& \textbf{Avg.} 

& \\

\midrule


{\color{blue}Gemini 3.1 Pro$^{\mathrm{T}}$} 

& \colorbox{cyan!20}{96.30} / \colorbox{cyan!20}{94.73} & \colorbox{cyan!20}{88.48} / \colorbox{cyan!20}{93.25} & \colorbox{cyan!20}{95.64} / \colorbox{cyan!20}{95.44} & \colorbox{cyan!20}{94.94} / \colorbox{cyan!20}{95.23} & \colorbox{cyan!20}{96.16} / \colorbox{cyan!20}{94.58} & \colorbox{cyan!20}{97.24} / \colorbox{cyan!20}{95.11} & \colorbox{cyan!20}{94.50} / \colorbox{cyan!20}{94.64} & 86.62 / \colorbox{cyan!20}{88.96} \\

{\color{blue}GPT-5.5$^{\mathrm{T}}$} 

& 96.07 / 94.69 & 88.13 / 93.08 & 95.40 / 95.35 & 94.58 / 94.89 & 95.93 / 94.47 & 97.06 / 95.08 & 94.23 / 94.51 & \colorbox{cyan!20}{86.66} / 88.72 \\

{\color{blue}GPT-5.5} 

& 95.93 / 94.56 & 88.26 / 92.48 & 95.37 / 95.24 & 94.49 / 94.78 & 95.83 / 94.31 & 97.01 / 94.81 & 94.20 / 94.28 & 86.51 / 88.00 \\


{\color{blue}DeepSeek-V4-Pro$^{\mathrm{T}}$} 

& 95.76 / 94.47 & 87.65 / 92.22 & 95.33 / {95.21} & 94.40 / 94.71 & 95.55 / 94.02 & 96.96 / 94.77 & 93.96 / 94.13 & 85.77 / 88.23 \\

{\color{blue}DeepSeek-V4-Pro} 

& 95.81 / 94.20 & 87.55 / 91.76 & 95.42 / 95.08 & 94.33 / 94.26 & 95.61 / 93.78 & 96.72 / 94.16 & 93.96 / 93.80 & 86.05 / 85.81 \\

{\color{blue}Kimi K2.6$^{\mathrm{T}}$} 

& 95.91 / 94.31 & 87.81 / 92.51 & 95.34 / 94.96 & 94.42 / 94.37 & 95.68 / 93.96 & 96.86 / 94.36 & 94.04 / 94.01 & 86.39 / 87.91 \\

{\color{blue}Kimi K2.6} 

& 95.30 / 93.74 & 87.30 / 91.07 & 94.95 / 94.42 & 93.98 / 93.68 & 95.18 / 93.43 & 96.66 / 93.75 & 93.58 / 93.27 & 85.88 / 87.09 \\

{\color{blue}GLM5.1$^{\mathrm{T}}$} 

& 95.77 / {94.55} & 87.72 / {92.54} & 95.08 / {95.21} & 94.40 / {94.87} & 95.60 / {94.29} & 96.82 / {94.86} & 93.92 / {94.30} & 86.10 / 88.29 \\

{\color{blue}GLM5.1} 

& 95.31 / 94.06 & 86.84 / 91.76 & 94.95 / 94.85 & 93.96 / 94.18 & 95.22 / 93.71 & 96.54 / 94.38 & 93.49 / 93.72 & 85.47 / 86.83 \\

{\color{blue}Qwen3.5-397B-A17B$^{\mathrm{T}}$} 

& 94.21 / 92.76 & 86.11 / 91.01 & 93.12 / 92.86 & 92.34 / 92.41 & 94.11 / 92.48 & 94.07 / 91.85 & 92.10 / 92.20 & 82.56 / 83.54 \\

{\color{blue}Qwen3.5-397B-A17B} 

& 95.88 / 94.32 & 87.59 / 91.63 & 95.24 / 94.94 & 94.38 / 94.35 & 95.64 / 93.80 & 97.13 / 94.60 & 93.98 / 93.82 & 86.97 / 87.56 \\

{\color{blue}Qwen3.6-35B-A3B$^{\mathrm{T}}$} 

& 96.03 / 94.31 & 87.58 / 91.66 & 95.35 / 94.87 & 94.53 / 94.23 & 95.76 / 93.76 & 96.91 / 94.33 & 94.06 / 93.77 & 87.24 / 87.75 \\

{\color{blue}Gemma4-31B$^{\mathrm{T}}$} 

& 96.07 / 94.32 & 87.66 / 92.13 & 95.37 / 94.89 & 94.15 / 94.27 & 95.89 / 93.80 & 97.12 / 94.55 & 94.07 / 93.90 & 86.87 / 87.33 \\

{\color{blue}Gemma4-26B-A4B$^{\mathrm{T}}$} 

& 95.78 / 94.30 & 87.44 / 91.81 & 95.19 / 94.83 & 94.11 / 94.13 & 95.50 / 93.68 & 96.92 / 94.29 & 93.84 / 93.76 & 86.30 / 86.03 \\

{\color{blue}Gemma4-E4B$^{\mathrm{T}}$} 

& 93.66 / 91.60 & 83.79 / 87.03 & 93.61 / 92.19 & 92.27 / 91.02 & 93.10 / 90.43 & 95.19 / 91.86 & 91.57 / 90.53 & 84.11 / 84.64 \\

{\color{blue}Gemma4-E2B$^{\mathrm{T}}$} 

& 92.77 / 91.25 & 82.69 / 86.77 & 93.12 / 91.67 & 91.94 / 90.56 & 92.41 / 89.77 & 94.94 / 91.25 & 90.90 / 90.05 & 83.12 / 82.85 \\

\hline

{\color{orange}Qwen3.6-35B-A3B} 

& 95.52 / 93.53 & 87.00 / 90.13 & 94.88 / 94.26 & 94.13 / 93.55 & 95.29 / 93.27 & 96.57 / 93.57 & 93.59 / 92.96 & 86.49 / 86.63 \\

{\color{orange}Gemma4-31B} 

& 95.56 / 93.81 & 87.10 / 91.28 & 95.12 / 94.44 & 94.02 / 93.97 & 95.28 / 93.14 & 96.77 / 94.17 & 93.65 / 93.34 & 86.35 / 86.55 \\

{\color{orange}Gemma4-26B-A4B} 

& 95.39 / 93.65 & 86.63 / 90.35 & 95.02 / 94.25 & 93.69 / 93.44 & 94.97 / 93.06 & 96.69 / 94.03 & 93.39 / 93.01 & 86.26 / 86.69 \\

{\color{orange}Gemma4-E4B} 

& 93.45 / 91.42 & 83.46 / 86.46 & 93.51 / 92.10 & 91.95 / 90.40 & 93.13 / 90.48 & 95.13 / 91.52 & 91.41 / 90.27 & 83.45 / 84.18 \\

{\color{orange}Gemma4-E2B} 

& 92.07 / 89.59 & 81.52 / 84.51 & 92.53 / 89.73 & 91.03 / 88.26 & 91.61 / 87.82 & 94.30 / 90.45 & 90.09 / 88.11 & 82.01 / 81.58 \\

{\color{purple}Tower-Plus-72B} 

& 95.23 / 93.60 & 86.54 / 90.80 & 95.04 / 94.35 & 93.91 / 93.62 & 95.41 / 93.29 & 96.53 / 93.83 & 93.46 / 93.14 & 85.13 / 86.66 \\

{\color{purple}TranslateGemma-27B-IT} 

& 96.33 / 92.38 & 87.56 / 88.88 & 95.81 / 93.84 & 94.66 / 92.29 & 96.00 / 92.01 & 97.26 / 93.20 & 94.30 / 91.94 & 88.44 / 85.65 \\

{\color{purple}Microsoft-Translator} 

& 92.51 / 90.34 & 81.76 / 85.17 & 92.47 / 90.64 & 91.19 / 87.96 & 92.77 / 90.25 & 94.96 / 90.88 & 90.49 / 89.01 & 79.19 / 79.32 \\

{\color{purple}Doubao-Translator} 

& 93.39 / 91.80 & 83.18 / 87.56 & 91.63 / 90.53 & 91.78 / 89.64 & 93.77 / 91.97 & 94.69 / 91.38 & 91.02 / 90.36 & 78.07 / 77.61 \\

\hdashline

\textbf{HY-MT1.5-1.8B} 
& 95.55 / 90.01 & 86.03 / 84.90 & 95.26 / 90.27 & 94.37 / 88.59 & 95.23 / 89.59 & 96.61 / 90.89 & 93.52 / 88.82 & 87.41 / 80.84 \\

\textbf{HY-MT1.5-7B} 

& 96.66 / 92.82 & 88.37 / 89.52 & \colorbox{orange!20}{96.28} / 93.40 & \colorbox{orange!20}{95.24} / 91.52 & 96.51 / 92.55 & \colorbox{orange!20}{97.57} / 93.09 & 94.82 / 92.04 & 90.21 / 87.13 \\

\hdashline

\textbf{Hy-MT2-1.8B} 

& 95.36 / 91.69 & 86.63 / 88.89 & 94.97 / 92.10 & 93.97 / 91.25 & 95.01 / 91.18 & 96.42 / 92.46 & 93.41 / 91.08 & 87.43 / 86.04 \\

\textbf{Hy-MT2-7B} 

& 96.79 / 93.14 & 89.02 / 91.14 & 96.23 / 93.76 & 95.15 / 92.80 & 96.47 / 93.02 & 97.50 / 93.43 & 94.92 / 92.79 & \colorbox{orange!20}{90.28} / 88.93 \\

\textbf{Hy-MT2-30B-A3B} 

& \colorbox{orange!20}{97.08} / \colorbox{orange!20}{94.14} & \colorbox{orange!20}{89.15} / \colorbox{orange!20}{92.04} & 96.22 / \colorbox{orange!20}{94.63} & 95.16 / \colorbox{orange!20}{94.08} & \colorbox{orange!20}{96.63} / \colorbox{orange!20}{93.79} & 97.54 / \colorbox{orange!20}{94.23} & \colorbox{orange!20}{95.04} / \colorbox{orange!20}{93.73} & 89.87 / \colorbox{orange!20}{89.25} \\

\bottomrule

\end{tabular}

}

\vspace{1mm}

\begin{minipage}{0.98\textwidth}

\tiny

\textit{Notes.} Each cell reports XCOMET / GEMBA scores, and all scores are multiplied by 100.
Avg. denotes the overall average score on DomainMTBench.
$^{\mathrm{T}}$ denotes thinking mode; for models with both modes, the row without $^{\mathrm{T}}$ denotes non-thinking mode.
Baselines are grouped into 
{\color{blue}large-scale models and all Think-mode models}, 
{\color{orange}medium to small-sized general models in non-Think mode}, and 
{\color{purple}translation-specialized models}. 
Our models are shown in bold.
The best results among large-scale models and all Think-mode models in each column are highlighted in \colorbox{cyan!20}{blue background}, while the best results among non-thinking-mode, small-to-medium-sized models are highlighted in \colorbox{orange!20}{light orange background}.
\end{minipage}

\end{table*}

\textbf{Domain-specific translation.} \ 
In domain-specific translation evaluation, we use DomainMTBench to assess model performance across multiple professional domains. We report XCOMET and GEMBA scores, where XCOMET measures reference-based translation quality and GEMBA provides an LLM-based assessment of overall translation quality. The results are shown in Table~\ref{tab:domain_wild_results}.

Overall, Hy-MT2 shows strong performance on DomainMTBench and consistently improves over Hy-MT1.5 in GEMBA. On the average score, Hy-MT2-1.8B improves the GEMBA score from 88.82 to 91.08 compared with Hy-MT1.5-1.8B, while Hy-MT2-7B improves from 92.04 to 92.79 compared with Hy-MT1.5-7B. Hy-MT2-30B-A3B further achieves 95.04 / 93.73 on the average score, obtaining the best XCOMET result among all compared systems and the best GEMBA result among open-source and translation-specialized models.
Across individual domains, Hy-MT2-30B-A3B achieves the best XCOMET scores in finance, law, and politics, with 97.08, 89.15, and 96.63, respectively. It also remains highly competitive in medical, technology, and education. These results indicate that the larger MoE model effectively strengthens domain-specific translation ability while maintaining strong performance across diverse professional domains. Hy-MT2-7B also demonstrates strong domain translation quality, improving over Hy-MT1.5-7B on the average score from 94.82 / 92.04 to 94.92 / 92.79, with clear gains on GEMBA across multiple domains. For the lightweight model, Hy-MT2-1.8B achieves a notable GEMBA improvement over Hy-MT1.5-1.8B, increasing the average score from 88.82 to 91.08. It also surpasses commercial systems such as Microsoft Translator and Doubao Translator on the average score, especially in GEMBA.

\textbf{In-the-wild translation.} \ 
On WildMTBench, Hy-MT2 also shows clear advantages in real-world translation scenarios. Hy-MT2-7B achieves 90.28 / 88.93, outperforming Hy-MT1.5-7B on both XCOMET and GEMBA. Hy-MT2-30B-A3B further reaches 89.87 / 89.25, obtaining the best GEMBA score among all compared systems and surpassing Gemini 3.1 Pro in LLM-based evaluation. For the lightweight setting, Hy-MT2-1.8B improves substantially over Hy-MT1.5-1.8B in GEMBA, increasing from 80.84 to 86.04, while maintaining a similar XCOMET score. It also clearly outperforms commercial translation systems such as Microsoft Translator and Doubao Translator on WildMTBench.

Overall, these results show that Hy-MT2 improves not only standard domain-specific translation, but also robustness and usability in real-world business scenarios. The consistent gains over Hy-MT1.5, especially in GEMBA, suggest that Hy-MT2 produces more natural and reliable translations under both professional-domain and in-the-wild settings.

\subsection{Instruction-Following Evaluation}

\begin{table*}[t]
\centering
\caption{Performance comparison on instruction-following benchmarks.}
\label{tab:if_results}
\scriptsize
\renewcommand{\arraystretch}{1.18}
\setlength{\tabcolsep}{3.2pt}
\resizebox{\textwidth}{!}{
\begin{tabular}{lccccccccc}
\toprule
\multirow{2}{*}{\textbf{Models}} 
& \multirow{2}{*}{\textbf{IFBench}} 
& \multirow{2}{*}{\textbf{IFEval}} 
& \multicolumn{3}{c}{\textbf{MaXIFE}} 
& \multirow{2}{*}{\textbf{Multi-IF}} 
& \multicolumn{3}{c}{\textbf{IFMTBench}} \\
\cmidrule(lr){4-6}
\cmidrule(lr){8-10}
& & & \textbf{Loose} & \textbf{Strict} & \textbf{Overall} & & \textbf{Simple} & \textbf{Complex} & \textbf{Total} \\
\midrule

{\color{blue}Gemini 3.1 Pro$^{\mathrm{T}}$}
& 71.33 & \colorbox{cyan!20}{96.30} & \colorbox{cyan!20}{90.58} & \colorbox{cyan!20}{87.62} & \colorbox{cyan!20}{89.10} & \colorbox{cyan!20}{95.02} / \colorbox{cyan!20}{89.57} / \colorbox{cyan!20}{84.89} & \colorbox{cyan!20}{91.95} & \colorbox{cyan!20}{84.53} & \colorbox{cyan!20}{89.08} \\

{\color{blue}GPT-5.5$^{\mathrm{T}}$}
& 67.33 & 93.90 & 89.96 & 84.58 & 87.27 & 94.90 / 89.33 / 84.46 & 89.25 & 84.44 & 87.39 \\

{\color{blue}GPT-5.5}
& 43.33 & 91.68 & 87.04 & 81.04 & 84.04 & 93.25 / 87.21 / 82.02 & 86.97 & 83.74 & 85.72 \\

{\color{blue}DeepSeek-V4-Pro$^{\mathrm{T}}$}
& \colorbox{cyan!20}{76.00} & 91.31 & 88.53 & 84.14 & 86.34 & 93.62 / 87.02 / 81.69 & 86.97 & 83.74 & 85.72 \\

{\color{blue}DeepSeek-V4-Pro}
& 42.67 & 89.83 & 84.98 & 78.77 & 81.88 & 90.36 / 83.45 / 76.60 & 81.38 & 78.23 & 80.16 \\

{\color{blue}Kimi K2.6$^{\mathrm{T}}$}
& 68.67 & 95.56 & 89.19 & 85.80 & 87.50 & 93.87 / 87.15 / 82.29 & 83.17 & 80.81 & 82.26 \\

{\color{blue}Kimi K2.6}
& 38.00 & 90.02 & 86.40 & 80.49 & 83.45 & 90.60 / 84.93 / 77.48 & 88.07 & 79.92 & 84.92 \\

{\color{blue}GLM5.1$^{\mathrm{T}}$}
& 74.00 & 94.09 & 88.59 & 85.61 & 87.10 & 93.94 / 86.47 / 82.77 & 88.81 & 82.14 & 86.23 \\

{\color{blue}GLM5.1}
& 50.00 & 90.57 & 86.40 & 80.49 & 83.45 & 92.24 / 85.25 / 79.67 & 85.53 & 78.83 & 82.94 \\

{\color{blue}Qwen3.5-397B-A17B$^{\mathrm{T}}$}
& 65.67 & 89.83 & 89.88 & 86.80 & 88.34 & 90.81 / 84.17 / 79.56 & 85.05 & 72.08 & 80.04 \\

{\color{blue}Qwen3.5-397B-A17B}
& 48.33 & 88.54 & 86.26 & 80.46 & 83.36 & 90.50 / 81.62 / 75.86 & 81.71 & 77.61 & 80.13 \\

\hline
{\color{orange}Gemma4-E2B}
& 26.00 & \colorbox{magenta!15}{80.44} & \colorbox{magenta!15}{77.32} & \colorbox{magenta!15}{68.79} & \colorbox{magenta!15}{73.06} & \colorbox{magenta!15}{80.79} / \colorbox{magenta!15}{70.59} / \colorbox{magenta!15}{63.81} & 62.93 & 49.10 & 57.59 \\

\textbf{Hy-MT2-1.8B}
& \colorbox{magenta!15}{35.33} & 80.22 & 61.29 & 51.05 & 56.17 & 70.50 / 49.79 / 35.75 & \colorbox{magenta!15}{76.76} & \colorbox{magenta!15}{57.61} & \colorbox{magenta!15}{69.36} \\

\hline
{\color{orange}Gemma4-E4B}
& 32.00 & 85.76 & \colorbox{magenta!15}{81.58} & \colorbox{magenta!15}{74.83} & \colorbox{magenta!15}{78.21} & \colorbox{magenta!15}{86.81} / \colorbox{magenta!15}{78.04} / \colorbox{magenta!15}{71.73} & 74.67 & 66.98 & 71.70 \\

\textbf{Hy-MT2-7B}
& \colorbox{magenta!15}{35.33} & \colorbox{magenta!15}{86.14} & 76.77 & 68.73 & 72.75 & 79.53 / 66.50 / 54.35 & \colorbox{magenta!15}{89.73} & \colorbox{magenta!15}{72.67} & \colorbox{magenta!15}{83.14} \\

\hline
{\color{orange}Qwen3.6-35B-A3B}
& 36.00 & 83.00 & 80.72 & 73.39 & 77.06 & 84.84 / 78.12 / 71.48 & 77.79 & 69.70 & 74.66 \\

{\color{orange}Gemma4-26B-A4B}
& 45.60 & \colorbox{magenta!15}{89.80} & \colorbox{magenta!15}{84.87} & \colorbox{magenta!15}{79.25} & \colorbox{magenta!15}{82.06} & 89.17 / \colorbox{magenta!15}{81.89} / \colorbox{magenta!15}{75.29} & 83.02 & 73.18 & 79.22 \\

\textbf{Hy-MT2-30B-A3B}
& \colorbox{magenta!15}{50.67} & \colorbox{magenta!15}{89.80} & 80.46 & 74.31 & 77.39 & \colorbox{magenta!15}{90.10} / 72.73 / 66.66 & \colorbox{magenta!15}{90.20} & \colorbox{magenta!15}{75.94} & \colorbox{magenta!15}{84.69} \\

\bottomrule
\end{tabular}
}
\vspace{1mm}

\begin{minipage}{0.98\textwidth}
\tiny
\textit{Notes.} All scores are reported as percentages. 
For Multi-IF, each cell reports turn1 / turn2 / turn3 scores. 
$^{\mathrm{T}}$ denotes thinking mode; for models with both modes, the row without $^{\mathrm{T}}$ denotes non-thinking mode.
IFMTBench is our translation instruction-following benchmark, and Simple, Complex, and Total correspond to single-constraint, multi-constraint, and overall scores, respectively.
Baselines are grouped into {\color{blue}ultra-large general models} and {\color{orange}medium to small-sized general models}. 
The best results within the ultra-large models are highlighted in \colorbox{cyan!20}{blue background}, while the best results within each corresponding parameter-scale group of small-to-medium models are highlighted in \colorbox{magenta!15}{pink background}.
\end{minipage}
\end{table*}





We evaluate instruction-following ability on both general instruction-following benchmarks and our translation-specific IFMTBench. The results are reported in Table~\ref{tab:if_results}.

Hy-MT2 shows strong translation-specific instruction-following ability on IFMTBench. Hy-MT2-7B achieves 89.73 / 72.67 / 83.14 on Simple, Complex, and Total, respectively, outperforming Gemma4-E4B and other medium-sized open-source baselines. Hy-MT2-30B-A3B further improves to 90.20 / 75.94 / 84.69, achieving the best overall IFMTBench score among small-to-medium-sized models. Compared with Qwen3.6-35B-A3B and Gemma4-26B-A4B, Hy-MT2-30B-A3B obtains consistent gains, especially on Complex instructions, showing stronger capability in handling multi-constraint translation requests.

The performance of Hy-MT2 is also competitive with much larger general-purpose models. On IFMTBench, Hy-MT2-30B-A3B approaches Kimi K2.6, GPT-5.5, and Gemini 3.1 Pro in Total score, and even surpasses several ultra-large models on Simple instructions. This indicates that targeted optimization for translation instruction following can effectively improve constraint understanding and execution.

On general instruction-following benchmarks, Hy-MT2-30B-A3B maintains solid performance, reaching 89.80 on IFEval and 77.39 overall on MaXIFE, outperforming Qwen3.6-35B-A3B on both metrics. However, its Multi-IF scores are lower in later turns than some comparable baselines, suggesting that the main advantage of Hy-MT2 lies in translation-specific instruction following rather than general multi-turn instruction following.

\subsection{Quantization Experiment}
We evaluate the model size and performance of various quantized Hy-MT2 models across general translation, domain-specific translation, and instruction-following benchmarks, as shown in Table~\ref{tab:quan_exp}. Overall, quantization substantially reduces deployment cost while preserving strong translation quality. For Hy-MT2-1.8B and Hy-MT2-7B, FP8 achieves performance very close to BF16 across most benchmarks, indicating that low-precision inference can be applied with minimal quality degradation. Q4\_K\_M also maintains competitive performance, especially on FLORES-200 and domain benchmarks, though larger drops appear on instruction-following tasks such as IFMTBench. The 2-bit Hy-MT2-1.8B shows a more noticeable decline, suggesting that extreme quantization introduces a stronger trade-off between efficiency and accuracy. For the MoE-based Hy-MT2-30B-A3B, FP8 remains highly stable compared with BF16, demonstrating that larger architectures are more robust to quantization. These results show that Hy-MT2 provides flexible deployment options, from highly efficient compact models to high-quality large models, while maintaining strong multilingual translation capability.

\begin{table}[t]
\centering
\caption{Performance comparison on various quantized model.}
\label{tab:quan_exp}
\setlength{\tabcolsep}{5.5pt}
\renewcommand{\arraystretch}{1.25}
\resizebox{\textwidth}{!}{
\begin{tabular}{lcccccccccc}
\toprule
\multirow{2}{*}{Model} 
& \multicolumn{3}{c}{FLORES-200} 
& \multirow{2}{*}{WMT25} 
& \multirow{2}{*}{Mand} 
& \multirow{2}{*}{DMTB} 
& \multirow{2}{*}{WMTB} 
& \multirow{2}{*}{IFBench} 
& \multirow{2}{*}{IFEVAL} 
& \multirow{2}{*}{IFMTB} \\
\cmidrule(lr){2-4}
& ZH $\Leftrightarrow$ XX 
& EN $\Leftrightarrow$ XX 
& XX $\Leftrightarrow$ XX 
& & & & & & & \\
\midrule
Hy-MT2-1.8B-BF16       & 83.49 & 87.02 & 79.21 & 60.33 & 60.32 & 92.25 & 86.74 & 35.33 & 80.22 & 69.36 \\
Hy-MT2-1.8B-FP8        & 83.11 & 86.66 & 78.63 & 59.51 & 59.73 & 92.15 & 86.57 & 35.00 & 80.59 & 67.06 \\
Hy-MT2-1.8B-Q4\_K\_M   & 82.22 & 85.87 & 77.19 & 57.46 & 57.23 & 91.85 & 86.09 & 33.33 & 78.93 & 63.47 \\
Hy-MT2-1.8B-2bit       & 80.86 & 84.74 & 76.31 & 57.96 & 58.96 & 89.67 & 83.33 & 34.00 & 77.63 & 58.99 \\
\hdashline
Hy-MT2-7B-BF16         & 89.17 & 91.43 & 87.06 & 73.05 & 65.49 & 93.86 & 89.61 & 35.33 & 86.14 & 83.14 \\
Hy-MT2-7B-FP8          & 88.92 & 91.45 & 86.92 & 72.59 & 65.50 & 93.84 & 89.49 & 34.00 & 86.51 & 82.38 \\
Hy-MT2-7B-Q4\_K\_M     & 88.96 & 91.46 & 86.90 & 72.30 & 65.57 & 93.81 & 89.41 & 37.67 & 85.58 & 75.11 \\
\hdashline
Hy-MT2-30B-A3B-BF16    & 90.04 & 92.37 & 88.13 & 73.62 & 65.93 & 94.39 & 89.56 & 50.67 & 89.80 & 84.69 \\
Hy-MT2-30B-A3B-FP8     & 89.97 & 91.87 & 88.01 & 73.38 & 65.91 & 94.35 & 89.61 & 50.33 & 88.70 & 83.47 \\
\bottomrule
\end{tabular}
}

\vspace{0.5em}
\begin{minipage}{0.98\textwidth}
\tiny
\textit{Notes.} All scores are reported as percentages. 
DMTB, WMTB, and IFMTB denote DomainMTBench, WildMTBench, and IFMTBench, respectively.
Mand denotes Mandarin-minority language bidirectional translation.
IFMTB is our translation instruction-following benchmark.
\end{minipage}
\end{table}

\subsection{Case Study on Translation Instruction Following} \label{sec:case_study}


\begin{figure}[!h]
\centering
\includegraphics[width=1.0\linewidth]{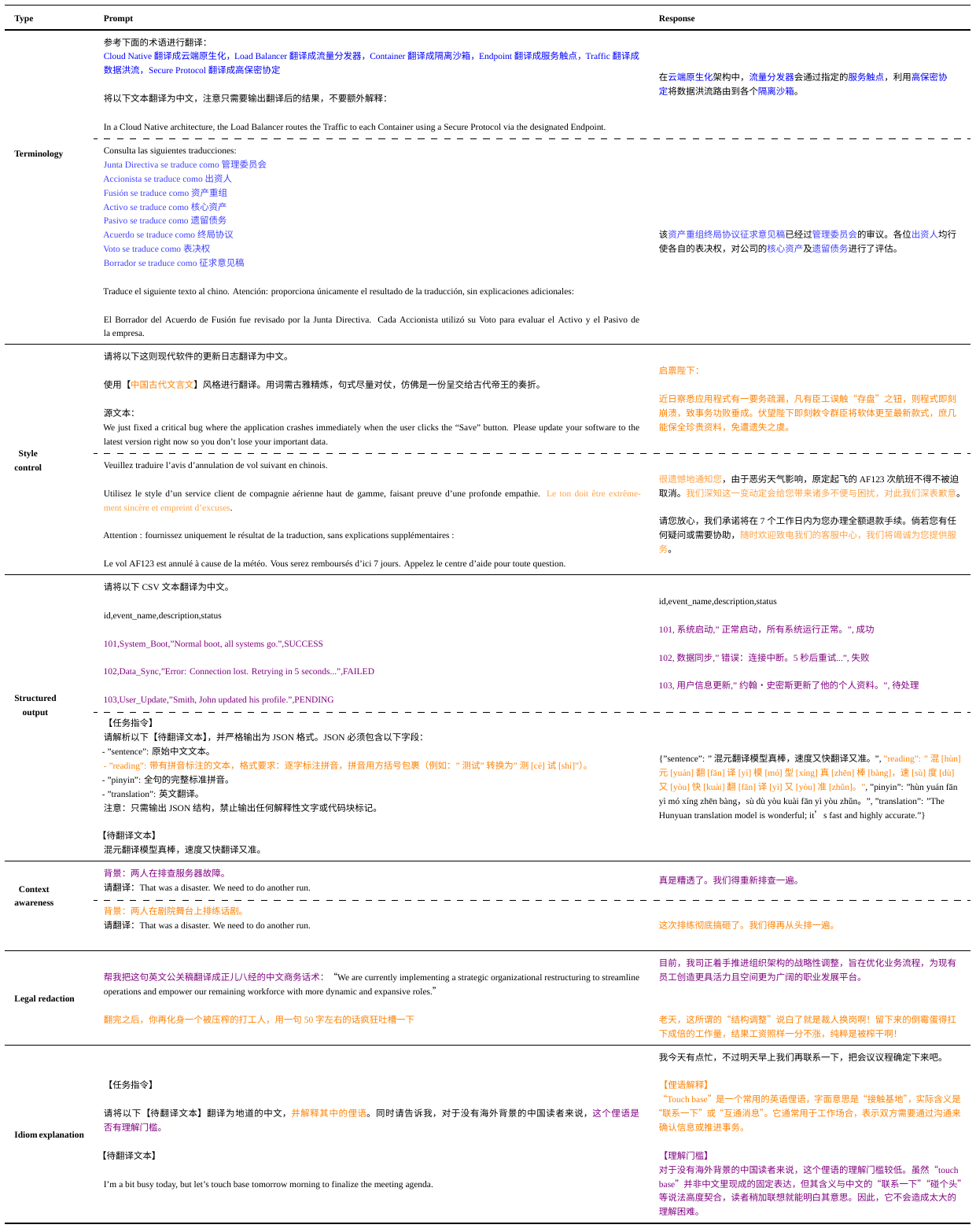}
\caption{Case study of Hy-MT2 on translation instruction following (Part 1). }
\label{fig:case_study1}
\end{figure}	

\begin{figure}[!h]
\centering
\includegraphics[width=1.0\linewidth]{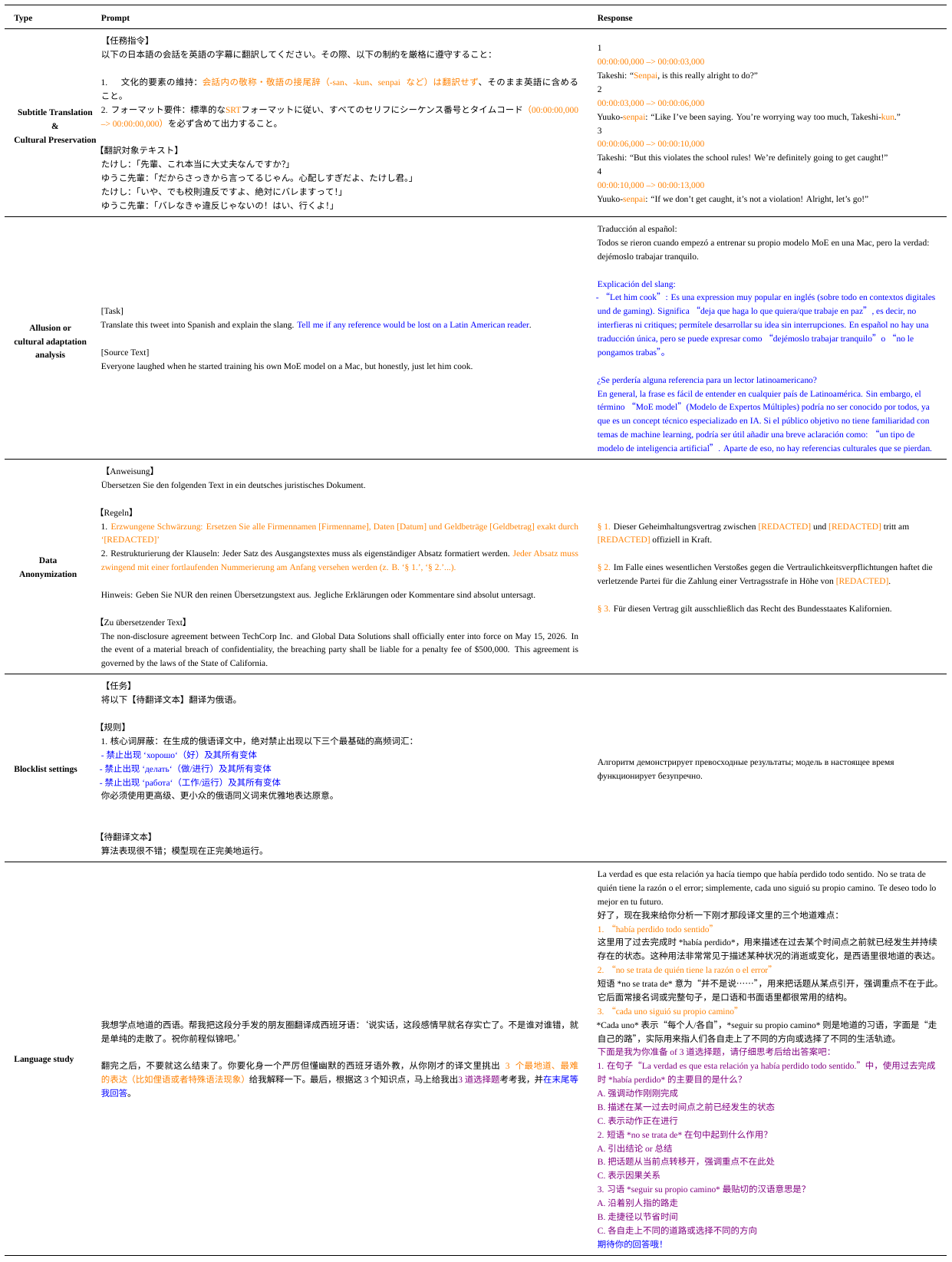}
\caption{Case study of Hy-MT2 on translation instruction following (Part 2). }
\label{fig:case_study2}
\end{figure}	

\begin{figure}[!h]
\centering
\includegraphics[width=1.0\linewidth]{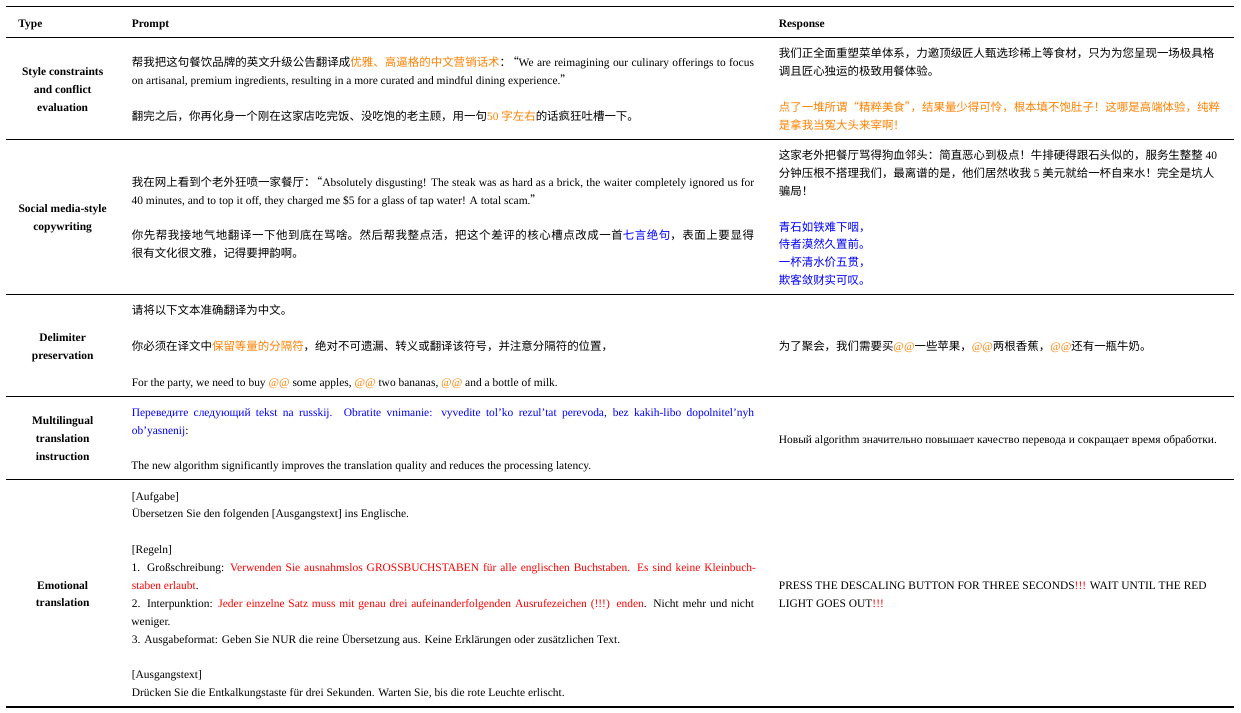}
\caption{Case study of Hy-MT2 on translation instruction following (Part 3). }
\label{fig:case_study3}
\end{figure}

Figures~\ref{fig:case_study1} to \ref{fig:case_study3} illustrate representative examples of Hy-MT2 handling translation-specific instructions. These cases encompass common real-world requirements, including style control, delimiter preservation, legal redaction, idiom explanation, subtitle translation with cultural preservation, legal-document anonymization, and language-learning-oriented annotation.

The examples demonstrate that Hy-MT2 can handle both explicit translation constraints and compound user requirements. In the style-control case, the model rewrites a modern software update log into classical Chinese with a formal, memorial-like tone. For delimiter preservation, it accurately maintains special markers (e.g., ``\#\#'') while translating surrounding content, showing precise formatting awareness.

Hy-MT2 also excels in more complex scenarios. In legal redaction and anonymization tasks, it translates while applying required redactions, paragraph restructuring, and numbering rules. In idiom-explanation and language-learning cases, it provides not only translations but also furigana, romaji output, and explanatory notes, meeting diverse instructional requirements.

Moreover, in subtitle translation, Hy-MT2 preserves culturally specific elements such as honorifics while following SRT formatting rules, reflecting its ability to manage multilingual instructions and maintain cultural fidelity.

Overall, these examples confirm that Hy-MT2 reliably executes diverse translation instructions, accommodating constraints on language, style, format, cultural context, and auxiliary explanatory needs.

\section{Conclusion}
In this paper, we present Hy-MT2, a multilingual machine translation model family designed for real-world translation scenarios. Hy-MT2 covers both dense and mixture-of-experts architectures, including \textbf{Hy-MT2-1.8B, Hy-MT2-7B, and Hy-MT2-30B-A3B}, all supporting translation among 33 languages. Compared with Hy-MT1.5, Hy-MT2 provides systematic improvements in domain-specific translation, real-world scenario translation, translation instruction following, model scaling, and efficient on-device deployment. Hy-MT2-7B and Hy-MT2-30B-A3B outperform strong open-source translation baselines such as DeepSeek-V4-Pro and Kimi K2.6, and achieve performance close to or even surpassing leading closed-source models such as Gemini 3.1 Pro on multiple benchmarks. The lightweight Hy-MT2-1.8B also demonstrates strong small-model translation capability, outperforming several commercial translation APIs. To support diverse deployment scenarios, Hy-MT2 is released in multiple precision formats, including \textbf{1.25-bit, 2-bit, 4-bit, 8-bit, and FP16}. Among them, the 1.25-bit and 2-bit versions are built on Hunyuan self-developed quantization techniques, significantly reducing model resource consumption while improving inference efficiency. Overall, Hy-MT2 provides a high-quality, efficient, and multi-capability multilingual translation model family for real-world applications.
\section{Contributions}

\subsection{Core Contributors}

Mao Zheng, Zheng Li, Tao Chen, Bo Lv, Mingrui Sun, Mingyang Song, Jinlong Song, Hong Huang, Decheng Wu, Hai Wang, Yifan Song, Yanfeng Chen, Guanwei Zhang

\subsection{Contributors}

Guanghua Yu, Yi Su, Hong Liu, Jinxiang Ou, Keyao Wang, Weile Chen, Haozhao Kuang, Kai Wang, Nuo Chen, Zihao Zheng, Chenhao Wang, Bin Xing, Chengcheng Xu, Tinghao Yu, Binghong Wu, Long Xu, Jiacheng Shi, Yunhao Wang, Baifang Chen, Lei Zhang, Qi Yang, Zhao Wu, Jiacheng Li, Lan Jiang, Lanrui Wang, Kai Zhang, Shuaipeng Li, Zhongzhi Chen, Weixuan Sun, Jiaqi Zhu, An Wang, Wei Li, Jun Xia, Weidong Han, Wutian Yang, Litong Hui, Luoguo Jia, Jiajia Wu, Hongchuan Zeng, Zheng Zhang, Xinpeng Zhou, Tianxiang Fei

\newpage

\end{CJK*}
\bibliography{colm2024_conference}
\bibliographystyle{colm2024_conference}
\end{document}